\documentclass[10pt,twocolumn,letterpaper]{article}

\usepackage{iccv}
\usepackage{times}
\usepackage{epsfig}
\usepackage{graphicx}
\usepackage{amsmath}
\usepackage{amssymb}
\usepackage{times}
\usepackage{epsfig}
\usepackage{booktabs}
\usepackage{multirow}
\usepackage{tabularx}
\usepackage{float}
\usepackage{color, colortbl}
\usepackage{xcolor}
\usepackage{adjustbox}
\usepackage{colortbl}
\usepackage{csquotes}
\usepackage{soul}
\usepackage{ctable} 
\usepackage{breqn}
\usepackage{authblk}

\usepackage[pagebackref=true,breaklinks=true,letterpaper=true,colorlinks,bookmarks=false]{hyperref}
\usepackage[normalem]{ulem}

\definecolor{Gray}{gray}{0.96}
\newcolumntype{a}{>{\columncolor{Gray}}c}

\newcommand{\bt}{\textbf}

\iccvfinalcopy 

\def\iccvPaperID{2628} 

\ificcvfinal\fi

\begin{document}

\title{P2C: Self-Supervised Point Cloud Completion from Single Partial Clouds}

\author{
Ruikai Cui$^{1}$\thanks{Email: ruikai.cui@anu.edu.au}\quad
Shi Qiu$^{1}\thanks{Corresponding author. Email: shi.qiu@anu.edu.au}$\quad
Saeed Anwar$^{2}$\quad
Jiawei Liu$^{1}$\quad
Chaoyue Xing$^{1}$\quad
Jing Zhang$^{1}$\quad
Nick Barnes$^{1}$
\\
$^{1}$Australian National University\quad
$^{2}$King Fahd University of Petroleum and Minerals
}

\maketitle
\ificcvfinal\thispagestyle{empty}\fi

\begin{abstract}
    Point cloud completion aims to recover the complete shape based on a partial observation. Existing methods require either complete point clouds or multiple partial observations of the same object for learning. In contrast to previous approaches, we present \textbf{Partial2Complete} (P2C), the first self-supervised framework that completes point cloud objects using training samples consisting of only a single incomplete point cloud per object. Specifically, our framework groups incomplete point clouds into local patches as input and predicts masked patches by learning prior information from different partial objects. We also propose Region-Aware Chamfer Distance to regularize shape mismatch without limiting completion capability, and devise the Normal Consistency Constraint to incorporate a local planarity assumption, encouraging the recovered shape surface to be continuous and complete. In this way, P2C no longer needs multiple observations or complete point clouds as ground truth. Instead, structural cues are learned from a category-specific dataset to complete partial point clouds of objects. We demonstrate the effectiveness of our approach on both synthetic ShapeNet data and real-world ScanNet data, showing that P2C produces comparable results to methods trained with complete shapes, and outperforms methods learned with multiple partial observations. Code is available at \url{https://github.com/CuiRuikai/Partial2Complete}.
\end{abstract}


\section{Introduction}
\label{sec:intro}

\begin{figure}[t]
\centering
\includegraphics[width=0.95\columnwidth]{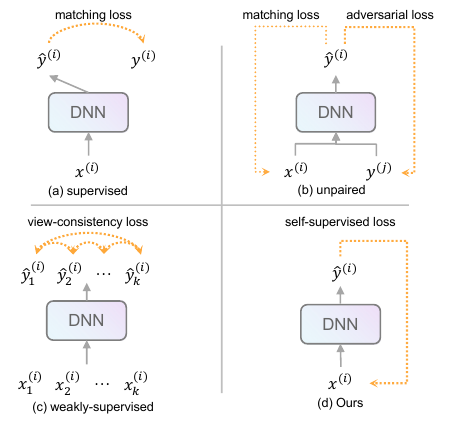}
\caption{Conceptual comparison of point cloud completion schemes.
Let $x_{k}^{(i)}$ be the $k$-th incomplete observation of object $i$, while $\hat{y}^{(i)}$ and $y^{(i)}$ be the corresponding completed prediction and ground truth, respectively. (a) Supervised approaches rely on paired partial-complete samples.
(b) Unpaired methods require partial point clouds and complete examples to guide predictions to match the input shape and follow the complete shape distribution.
(c) Weakly-supervised models learn completion based on consistency across multi-view partial samples of an object.
(d) Our scheme differs from existing settings as only a single partial observation per object instance is available for learning.}
\label{fig:completion}
\end{figure}

Point clouds are widely used for 3D shape representation and play a crucial role in a range of applications~\cite{poisson_Kazhdan, pnp3d_qiu, qiu2021geometric, qiu2021semantic}.
However, real-world raw point clouds are collected from sources such as laser scanners~\cite{kitti_geiger} and depth cameras~\cite{scannet_dai}, and so are often incomplete and noisy due to occlusions and varying lighting conditions.
For this reason, point cloud completion (PCC)~\cite{pcn_yuan, epn_dai, topnet_Tchapm,high_han} is studied to obtain complete point clouds from partial ones.

Supervised learning~\cite{pointr_yu, topnet_Tchapm, vrcnet_pan} offers a straightforward solution, where both partial point clouds and ground truth completions are required during training.
Nevertheless, collecting complete point clouds is challenging.
As a result, training data pairs are often obtained by simulating occlusions on 3D model collections like ShapeNet~\cite{shapenet_chang}.
Due to the distribution gap between real and simulated data, the real-world performance of these approaches  is often limited.

Unpaired (or unsupervised) PCC~\cite{pcl2pcl_chen} is an alternative to supervised PCC, which trains a category-specific network using only partial point clouds and a set of example complete shapes of the same category.
This approach enables the use of incomplete shapes from large-scale real scans and virtual 3D object datasets, as the partial points and complete shapes
do not need to be
paired.
However, obtaining a large, complete, and clean 3D point cloud dataset remains challenging, due to factors such as labor cost, equipment expenses, \emph{etc}.
Weakly-supervised methods~\cite{wild_gu,pointpncnet_mittal,traffic_ren} have been proposed by constructing weak supervision cues using multiple unaligned observations from different views of the same object.
However, the performance can be significantly affected by alignment errors, and collecting observations from many views is difficult due to hardware limitations or viewing angle restrictions.

To address these challenges, we propose a new self-supervised approach to PCC, where for training, we only require one point cloud observation with unknown incompleteness per object.
This novel setting offers several benefits for completion:
1) it eliminates the need for complete samples, thereby reducing the difficulty and expense of annotation;
2) partial objects can be easily collected from the actual world even if only a single viewing angle is available, significantly expanding the scope of training data;
3) by leveraging the unknown incompleteness assumption, partial samples, complete shapes and weakly-supervised cues can be unified in the learning framework to improve completion quality.
Fig.~\ref{fig:completion} illustrates the difference of our proposed setting with existing main schemes.

In this paper, we introduce Partial2Complete (P2C), an effective approach for training a category-specific point cloud completion network using only single partial point clouds.
Inspired by He \emph{et al.}~\cite{masked_he}, P2C groups input points as patches that represent a small but possibly continuous region on the underlying surface, where we expect the network to predict masked patches based on unmasked regions.
Our approach assumes that a structural prior can be learned by observing a number of training objects with different missing parts, guiding the reconstruction of severely incomplete point clouds.
Furthermore, we develop the cycle constraint~\cite{cyclegan_zhu} from unpaired image translation to propose a latent reconstruction loss to the framework. This regularization ensures that completing different partial regions of the same object leads to the same completed shape.

We also present two new components to address problems that are unique to the self-supervised setting.
First, traditional point cloud distance measures~\cite{pcn_yuan, cycle_wen} lack awareness of complete or missing regions that occur in the completion task, leading to either limited completion capability or mismatching predictions.
To address this challenge, we introduce Region-Aware Chamfer Distance (RCD) to estimate point cloud correspondence based on regions centered at dynamically generated skeleton points. By optimising RCD, possible outlier points can be pulled to the target point set and completion of missing regions will not be restricted.
On the other hand, motivated by techniques that use differential geometry-based surface curvature to describe and identify local surface shape~\cite{caeli,liang_todhunter, unique_tombari, histogram_tang}, we propose  the Normal Consistency Constraint (NCC) to encourage generated points to follow the local 2D surface manifold of the incomplete point cloud.
The NCC queries the normal direction similarity for nearby points and computes the similarity variance as a regularizer to encourage local planarity.

We apply P2C to synthetic and real-world completion tasks to comprehensively verify its effectiveness.
We show that, without any complete shape examples, our approach not only achieves comparable results against methods with access to complete samples, but also outperforms weakly-supervised methods trained with multiple incomplete observations.
In summary, our main contributions are:
\begin{itemize}
\item We propose, P2C, the first self-supervised framework that is able to complete point clouds with only a single partial point cloud per object for learning.
\item We design a novel distance measure, Region-Aware Chamfer Distance, which
overcomes problems of restricting completion and insufficient supervision, by constructing local regions around dynamically constructed skeleton points.
\item We present the Normal Consistency Constraint to refine shape predictions to follow the local surface manifold by minimizing a novel consistency metric, improving surface continuity and completeness.
\end{itemize}

\begin{figure*}
\centering
\includegraphics[width=0.98\textwidth]{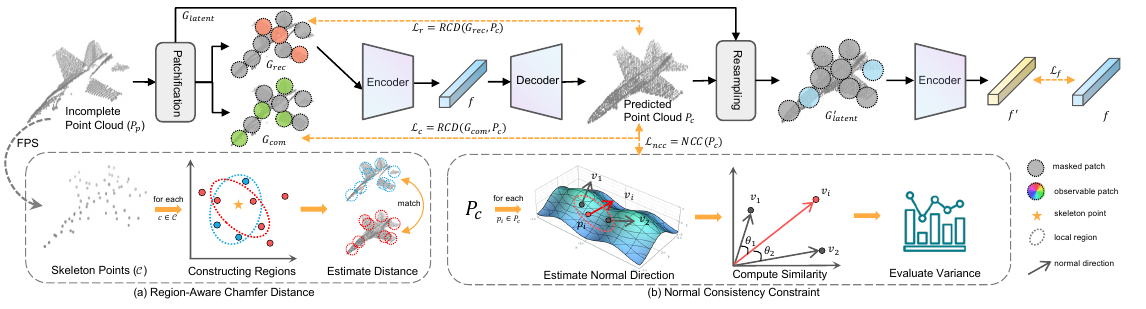}
\caption{The Pipeline of \textit{P2C}.
Starting from the partial point cloud $P_p$, we divide it into patches and partition these patches into three groups ($G_{rec}, G_{com}, G_{latent}$).
The encoder takes $G_{rec}$ to produce features $f$ then the decoder generates a predicted point cloud $P_c$ based on $f$. $G_{latent}$ is never observed by the encoder, we resample corresponding regions $G_{latent}^{'}$ in $P_c$ to yield another feature embedding $f'$.
The overall loss has four components.
The reconstruction loss $\mathcal{L}_{r}$ and completion loss $\mathcal{L}_{c}$ are realized by RCD.
The Latent reconstruction loss $\mathcal{L}_{f}$ and the normal consistency constraint $\mathcal{L}_{ncc}$ are introduced to regularize the inference.}
\label{fig:pipeline}
\end{figure*}

\section{Related Works}
\label{sec:related}

\noindent\textbf{Supervised Point Cloud Completion.}
Earlier efforts to address point cloud completion can be divided into surface reconstruction and template matching.
Surface reconstruction methods~\cite{poisson_Kazhdan, screen_Kazhdan} attempt to restore missing regions by fitting existing points to an implicit surface based on geometric cues, and then resample new points from the estimated surface.
On the other hand, template matching techniques~\cite{data_sung, example_pauly} retrieve a template shape from a database and deform it to fit the target shape.
However, surface reconstruction-based methods are able to fill holes on the surface but are limited in handling severe geometric incompleteness, while template matching methods are computationally expensive and rely on the availability of a sufficient number of example shapes.
Starting with the pioneering work PCN~\cite{pcn_yuan}, deep learning-based methods~\cite{folding_yang, snowflake_xiang, vrcnet_pan, pointr_yu} have gained significant attention in point cloud completion.
However, the supervised training approach requires paired ground truth, which is difficult to obtain for real-world scans.
As a result, these methods are often trained on synthetic datasets, which leads to impressive results on synthetic data but may not generalize well to real-world scans~\cite{shapeinver_zhang}.

\noindent\textbf{Unpaired and Weakly-Supervised Completion.}
To address the issue of data acquisition, Chen~\emph{et al.}~\cite{pcl2pcl_chen} proposed the first method, Pcl2Pcl, that can be trained without paired partial and complete point sets.
This was achieved through a generative adversarial network~\cite{gan_goodfellow}, where the generator transforms a partial shape latent encoding into a representation indistinguishable from the latent variable obtained from real complete shapes by the discriminator.
Following Pcl2Pcl, many methods~\cite{cycle_wen, shapeinver_zhang, latent_cai,ebm_cui} have been proposed to produce more accurate results.
Nevertheless, complete shape repositories are still required, and combining unaligned real-world partial scans with complete shapes from other sources may result in poor outcomes due to alignment errors.
Different from prior approaches, Gu \emph{et al.}~\cite{wild_gu} tackle the problem of point cloud completion by using unaligned real-world partial point clouds as their data source.
The network is trained with multi-view geometric constraints as weak supervision cues.
However, these methods require scans from multiple viewing angles, which are not always feasible to obtain.

\noindent\textbf{Self-Supervised Learning.}
To mitigate the cost of dataset collection and annotation, self-supervised learning~\cite{sl_survey_goyal} have been proposed.
For example, DINO~\cite{dino_caro} demonstrated improved classification performance using only self-supervised training, without any labels.
Self-supervised learning has also gained popularity in point cloud studies.
Building upon the work of He~\emph{et al.}~\cite{masked_he}, Liu~\emph{et al.}~\cite{masked_liu} proposed a self-supervised mask discrimination framework for pretraining transformers.
For point cloud upsampling, SSPU-Net~\cite{sspu_zhao} leverages the consistency between input sparse and generated dense point clouds to train the network using only sparse clouds.
Concurrently with our research, Hong \emph{et al}.~\cite{acl_hong} proposed a related point cloud completion scheme, but used the same data for training and testing to enable an adaptive closed-loop~\cite{acl_astrom} optimization. In contrast, our approach uses distinct test samples.

\section{Method}

A complete point cloud can be generated by uniformly sampling an underlying object surface, while an incomplete point cloud is obtained from the surface via biased sampling, \eg~due to occlusion.
Our proposed self-supervised point cloud completion method aims to predict an object's complete shape, given only a single incomplete point cloud per object from the same object category during learning.
The key motivation of our method is to recover the missing part of one object by observing similar regions of other objects in the same category.
Accordingly, even if a large shape collection contains only partial objects, as long as all kinds of parts of a category are exhibited across multiple object instances, the dataset is sufficient for learning to complete partial shapes.

Given only partial observations, our method learns completion via patch-wise self-supervised learning (Sec.~\ref{sec:p2c}), where patches (Sec.~\ref{sec:patch}) of the partial point cloud are generated to achieve both shape augmentation and region-aware regularization (Sec.~\ref{sec:rcd}).
Further, we introduce the Normal Consistency Constraint during training (Sec.~\ref{sec:ncc}) to enforce the assumption that object surfaces are continuous and closed by leveraging a local planarity along the object surface.
The overall pipeline is depicted in Fig.~\ref{fig:pipeline}.

\subsection{Partial2Complete}
\label{sec:p2c}

Let $P_p$ be an incomplete point cloud and $P_c$ a predicted completion of $P_p$.
Our framework takes $P_p$ as input to generate $M$ patches, each of which represents a local region on the surface of the observed shape.
The $M$ patches are partitioned into three groups $\{G_{rec}, G_{com}, G_{latent}\}$.
$G_{rec}$ is the observable region for the network, and we force the network to generate a shape prediction $P_c$ that preserves the regions in $G_{rec}$ by introducing the reconstruction loss $\mathcal{L}_r$.
Although $\mathcal{L}_r$ effectively regularizes the predicted shape to match the observed regions in $G_{rec}$, this loss alone is not enough to guide the network to predict a complete shape.
To this end, the completion loss $\mathcal{L}_c$ is used to penalize the network for not predicting the masked group $G_{com}$.
Manually masked parts and those missing from the input are both unseen by the network, hence, are indistinguishable for the network, and so minimizing $\mathcal{L}_c$ guides the network to complete both naturally absent and intentionally masked regions.

The first group $G_{rec}$ is passed through the encoder to obtain a latent feature embedding $f$, representing an encoding of the corresponding object and serving as input to the decoder to produce a shape prediction $P_c$. To further regularize the completion, we introduce latent reconstruction loss $\mathcal{L}_{f}$ to encourage two different sets of local regions of an object sharing the same object latent representation~\cite{cyclegan_zhu}.
Particularly, we exploit the third set of patches $G_{latent}$, which is separate from $G_{rec}$ and not observed by the encoder. By resampling $G_{latent}$ in $P_c$, we collect the patches at the same spatial location as another group $G_{latent}^{'}$.
Then, we pass $G_{latent}^{'}$ to the encoder, resulting in a latent feature $f'$, and $\mathcal{L}_{f}$ is utilized to penalize the difference between $f$ and $f'$.

\subsection{Patchification and Partition}
\label{sec:patch}

We sample patches from the object surface to provide information about local regions. To achieve this, we use farthest point sampling (FPS)~\cite{pointnet2_qi} to sample $M$ points as patch centers $C=\{c_i\}_{i=1}^{M}$ from partial shape $P_p$.
Then, we gather the $k$-nearest neighbors of each center point based on Euclidean distance to obtain a patch $g_i=\{p| p \in \mathcal{N}_{k}^{P_p}(c_i)\}$ where $\mathcal{N}_{k}^{P_p}(c_i)$ denotes the set of $k$-nearest neighbors for $c_i$ in $P_p$.
Furthermore, the patches are divided into the three partitions: $G_{rec}$, $G_{com}$, and $G_{latent}$, with ratio $r_1:r_2:r_3$.
Once the decoder produces the predicted shape $P_c$, we resample $G_{latent}^{'}$ as the regions corresponding to $G_{latent}$ in the prediction by employing the same patch centers used for $G_{latent}$ and searching for the k-nearest neighbors in $P_c$.

\subsection{Region-Aware Chamfer Distance}
\label{sec:rcd}

\begin{figure}
\centering
\includegraphics[width=\columnwidth]{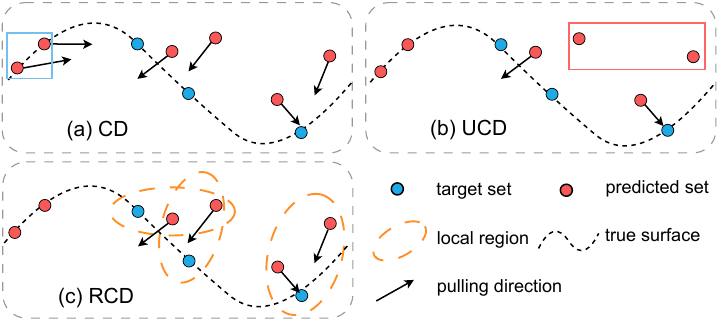}
\caption{Comparison of pulling direction to minimize different distance measures. (a) CD takes the nearest neighbor for every point in the predicted set, leading to restrictions in completing missing parts;
(b) UCD considers the nearest neighbor only for every point in the target set, resulting in no moving directions for noisy points;
(c) RCD is aware of observed and unseen regions and thus only evaluates point distance for observed regions, pulling outlier points to the underlying surface while allowing completion of unseen parts.
}
\label{fig:rcd}
\end{figure}

Chamfer Distance (CD) and Unidirectional Chamfer Distance (UCD) are commonly used to measure the distance between two point clouds that may have different numbers of points~\cite{pcn_yuan, cycle_wen}.
UCD between two point sets $S_1$ and $S_2$ is defined as follows:
\begin{equation}
\label{eq:cd}
d_{UCD}(S_1, S_2) = \frac{1}{|S_1|}\sum_{x\in S_1}\min_{y\in S_2}\|x-y\|_2.
\end{equation}
CD takes both directions into account and can be defined through UCD as $d_{CD} = d_{UCD}(S_1, S_2) + d_{UCD}(S_2, S_1)$.

Let $P_p$ be a partial point cloud of an object with some missing regions, and $P_c$ be a prediction corresponding to a complete but possibly noisy shape of the same object.
When applying the two distance measures to self-supervised completion, where we have no access to a complete shape as ground truth, CD is not aware of incompleteness while UCD has no regularization for outliers.
For $d_{CD}(P_p, P_c)$, predicted points $p\in P_c$ that correspond to unseen parts in the partial shape are estimated as far away from the underlying surface.
Therefore, as shown in Fig.~\ref{fig:rcd}~(a), the two points in the blue box are located on the true surface of the object, but they will be displaced to minimize the CD.
Thus, CD prevents the network from inferring missing parts.
Moreover, $d_{UCD}(P_p, P_c)$ measures the distance by only considering the points in the prediction that are the nearest neighbors of points in $P_p$.
We show the effect on UCD in Fig.~\ref{fig:rcd}~(b) that although completion of unseen regions will not be restricted, outlier points bounded by the red box are less likely to be selected as nearest neighbors of points in the target set, leading to no distance measure for outliers in the prediction.
As a consequence, the network will not learn to avoid outliers in the prediction when using UCD as the distance measure.

Region-Aware Chamfer Distance (RCD) addresses the problem of seen/unseen region awareness by constructing local regions in both prediction and partial input centered at skeleton points that are dynamically sampled from the partial shape $P_p$.
Specifically, given two point sets, $P_p$ and $P_c$, $m$ points are sampled from $P_p$ as skeleton points $\mathcal{C} = \{c_i\}_{i=1}^m$ through farthest point sampling~\cite{poinnet2_qi}, representing a rough observed shape.
Then, the $k$-nearest neighbors in each point set are gathered for each skeleton point in $\mathcal{C}$, forming two sets that represent the matched regions $R_p$ and $R_c$.
Then, RCD can be defined through UCD as:
\begin{equation}
    \label{eq:rcd}
    d_{RCD}(P_p, P_c) = d_{UCD}(R_p, R_c) + d_{UCD}(R_c, R_p),
\end{equation}
where
\begin{align*}
R_{p} & = \bigcup_{i=1}^{m}\{\mathcal{N}_k^{P_p}(c_i)| \ c_i \in \mathcal{C}\} \text{ for } P_p,\\
R_{c} & = \bigcup_{i=1}^{m}\{\mathcal{N}_k^{P_c}(c_i)| \ c_i \in \mathcal{C}\}  \text{ for } P_c,
\end{align*}
are the union of $k$-nearest neighbors for all skeleton points in $P_p$ and $P_c$, respectively.

\subsection{Normal Consistency Constraint}
\label{sec:ncc}

\begin{figure}
\centering
\includegraphics[width=\columnwidth]{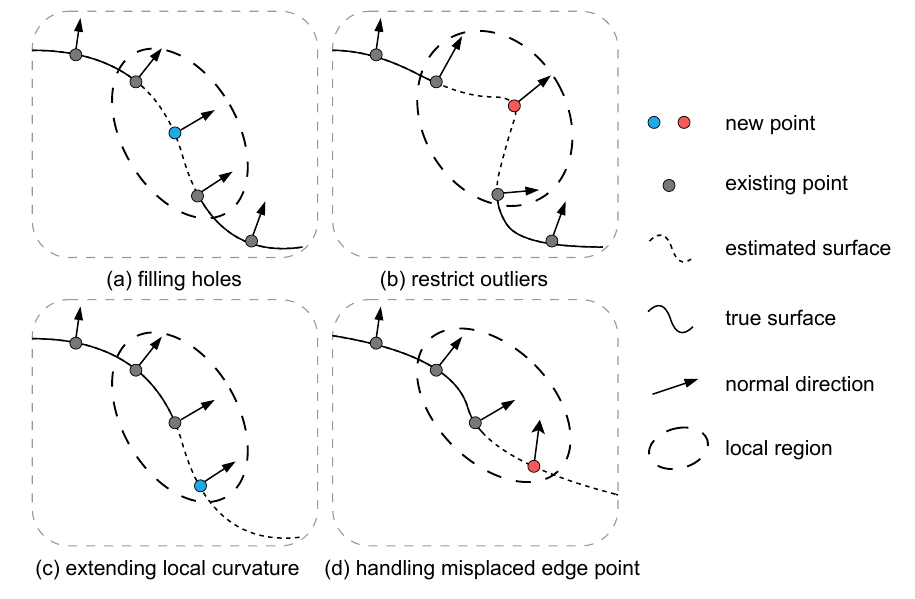}
\caption{Illustration of the effect of NCC in a 2D case: The variance of normal similarity is lower when the point follows the underlying surface, as shown in (a) and (c), while the variance is larger when the new point results in a surface that diverges from the existing surface curvature, as shown in (b) and (d).}
\label{fig:ncc}
\end{figure}

To further regularize the completion, we introduce the Normal Consistency Constraint (NCC) to improve surface continuity.
Specifically, given a point cloud $P=\{p_i\}_{i=1}^{n}$, the total least squares estimation of the normal direction~\cite{caeli} of a tangent plane centered at $p_i$ is obtained by eigenvalue decomposition of the covariance matrix $Cov$ of the $k$-nearest neighbors $\forall q_j\in\mathcal{N}_k^{P}(p_i)$, defined as:
\begin{equation}
    \label{eq:covariance}
    Cov = \frac{1}{k}\sum_{j=1}^{k}{(q_j-\hat{p})(q_j-\hat{p})^T},\ \hat{p}=\frac{1}{k}\sum_{j=1}^{k}{q_j},
\end{equation}
where the eigenvector corresponding to the smallest eigenvalue of $Cov$ is the estimated normal direction $v_i$, and $v_i$ is normalized as $\|v_i\|=1$.
We define the normal consistency of a point $p_i$ as:
\begin{equation}
\label{eq:nc}
nc(p_i)= \bigg(\sum_{j=1}^k (v_i^Tv_j - \mu_i)^T(v_i^Tv_j - \mu_i)\bigg)^{{1}/{2}},
\end{equation}
where a dot product between two normal directions is applied as the similarity measure, and $\mu_i = \frac{1}{k}\sum_{j=1}^{k}{v_i^Tv_j}$ is the mean of similarities between $v_i$ and $v_j$.
The value of $nc(\cdot)$ represents the variance of the normal similarity, which estimates the local surface curvature.
As the local surface approaches piece-wise planar, $nc(\cdot)$ decreases to 0, while $nc(\cdot)$ increases as the curvature increases. Further, NCC is formulated as:
\begin{equation}
    \label{eq:ncc}
    NCC(P)=\frac{1}{n}\sum_{i=1}^{n}nc(p_i).
\end{equation}

As illustrated in Fig.~\ref{fig:ncc},
when a new point added to fill a hole or extend an edge following the local plane, this results in a smaller $nc$ value than if the point diverges from the surface curvature.
Therefore, the NCC regularizes the prediction to be more smooth and extends edge points to make the prediction more complete, leading to better shape completion.

\subsection{Optimization}
\label{sec:loss}

The reconstruction loss and completion loss are defined as $\mathcal{L}_r = d_{RCD}(G_{rec}, P_c)$ and $\mathcal{L}_c = d_{RCD}(G_{com}, P_c)$, respectively.
We encode $G_{latent}^{'}$ in $P_c$ as a latent representation $f' \in \mathbb{R}^d$, and we encourage it to be consistent with the first latent embedding $f\in \mathbb{R}^d$ via latent reconstruction loss:
\begin{equation}
    \label{eq:latent_recon_loss}
    \mathcal{L}_{f}(f, f') = \frac{1}{d}\sum_{i=1}^{d} \phi(f_i-{f_i}'),
\end{equation}
where $\phi(\cdot)$ is the Huber~\cite{huber_huber} loss function.
Together with the NCC as a loss function $\mathcal{L}_{ncc} = NCC(P_{c})$, we have the overall loss defined as:
\begin{equation}
    \mathcal{L} = \lambda_{rec}\mathcal{L}_{r} + \lambda_{com}\mathcal{L}_{c} +  \lambda_{latent}\mathcal{L}_{f} +  \lambda_{ncc}\mathcal{L}_{ncc},
\end{equation}
where $\lambda_{rec}, \lambda_{com}, \lambda_{latent}, \lambda_{ncc}$ are weighting parameters.

\section{Experiments}

\subsection{Implementation Details}
\label{sec:implementation}
We employ the encoder from PCN~\cite{pcn_yuan} for our method. The decoder is implemented as a multi-layer perceptron with two hidden layers of 2048 dimensions.
For the loss functions, we set the weights for the reconstruction loss, completion loss, and latent reconstruction loss to 1, 1, and 0.1, respectively.
The weight for the NCC loss is set to 0.1. The number of patches used is 64, each formed by a local region of 32 points.
The three groups $G_{rec}, G_{com}, G_{latent}$ each contain 20, 40, and 4 patches, respectively.
The network is trained using the AdamW~\cite{adamw_Loshchilov} optimizer with a starting learning rate of $10^{-3}$ and a weight decay of $10^{-3}$ for $300$ epochs.

\subsection{Dataset and Evaluation Metrics}

\begin{table*}
\centering
\caption{Quantitative comparison result of our method and other methods on the 3D-EPN dataset using CD-$\ell_2$ $\downarrow$ ($\times 10^4$).}
\begin{tabular}{l|l|a|cccccccc}
\toprule
Method                               & Data Source  & Average & Plane & Cabinet & Car  & Chair & Lamp & Couch & Table & Boat \\ \midrule
FoldingNet~\cite{folding_yang}       & paired       & 6.8   & 2.6   & 7.6     & 4.8  & 8.3   & 9.7  & 7.4   & 8.0   & 5.8     \\
PCN~\cite{pcn_yuan}                  & paired       & 7.4   & 2.5   & 8.0     & 4.8  & 9.0   & 12.2 & 8.1   & 8.9   & 6.0     \\
TopNet~\cite{topnet_Tchapm}          & paired       & 6.4   & 2.3   & 7.5     & 4.6  & 7.6   & 8.9  & 7.3   & 7.5   & 5.2       \\ 
PoinTr~\cite{pointr_yu}              & paired       & 4.3   & 1.2   & 6.5     & 4.0  & 5.1   & 4.5  & 5.4   & 5.4   & 2.6   \\ \midrule
Pcl2Pcl~\cite{pcl2pcl_chen}          & unpaired     & 17.4  & 4.0   & 19.0    & 10.0  & 20.0  & 23.0 & 26.0  & 26.0  & 11.0   \\
C4C~\cite{cycle_wen}                 & unpaired     & 14.3  & 3.7   & 12.6    & 8.1   & 14.6  & 18.2 & 26.2  & 22.5  & 8.7      \\
Inv~\cite{shapeinver_zhang}          & complete & 23.6  & 4.3   & 20.7    & 11.9  & 20.6  & 25.9 & 54.8  & 38.0  & 12.8     \\
Cai \emph{et al}.~\cite{latent_cai}  & unpaired     & 13.6  & \bt{3.5} & \bt{12.2} & 9.0 & 12.1 & 17.6 & 26.0 & 19.8 & 13.6     \\
P2C$^*$(Ours)                           & unpaired     & \textbf{10.9} & 3.7 & 12.5 & \textbf{7.7} & \textbf{11.3} & \textbf{15.3} & \textbf{13.2} & \textbf{15.2} & \textbf{8.0}  \\ \midrule
Gu \emph{et al.}~\cite{wild_gu}      & multi-view   & 21.3  & 5.9   & 20.8    & 9.5  & 20.4  & 34.9 & 27.1  & 36.7  & 14.8  \\ 
PPNet~\cite{pointpncnet_mittal}      & multi-view   & 28.1  & 5.6   & 46.6    & 22.4 & 24.3  & 46.1 & 28.4  & 36.4  & 15.0  \\ 
P2C(Ours)                            & single partial & \textbf{14.1} & \textbf{4.3} & \textbf{19.4} & \textbf{8.6} & \textbf{13.5} & \textbf{16.3} & \textbf{20.2} & \textbf{18.1} & \textbf{12.0} \\ 
\bottomrule
\end{tabular}
\label{tb:epn}
\end{table*}

\begin{table}
\caption{Quantitative comparison result of our method and other methods on the PCN dataset using CD-$\ell_2$ $\downarrow$ ($\times 10^4$).}
\begin{adjustbox}{width=\columnwidth, center}
\setlength\tabcolsep{2pt}
\begin{tabular}{l|cccccccc|a}
\toprule
Method                          & Air   & Cab  & Car  & Cha   & Lam   & Sof   & Tab   & Wat  & Avg  \\ \midrule
Folding~\cite{folding_yang}     & 2.4   & 8.4  & 4.9  & 9.2   & 11.5  & 9.6   & 8.4   & 7.4  & 7.7  \\
PCN~\cite{pcn_yuan}             & 3.0   & 7.5  & 5.7  & 9.7   & 9.2   & 9.5   & 9.2   & 6.2  & 7.5  \\
TopNet~\cite{topnet_Tchapm}     & 2.3   & 8.2  & 4.7  & 8.6   & 11.0  & 9.3   & 7.5   & 5.2  & 6.4  \\ \midrule
C4C~\cite{cycle_wen}            & 4.1   & 14.2 & 9.9  & 14.6  & 19.2  & 27.8  & \bt{16.8}  & 9.0  & 14.4   \\
Inv~\cite{shapeinver_zhang}    & 3.9   & 17.4 & 11.0 & 13.8  & \bt{14.2}  & 23.0  & 20.3  & 9.7  & 14.1   \\ 
P2C(Ours)                       & \bt{3.5} & \bt{11.7} & \bt{9.0} & \bt{12.8} & 16.4 & \bt{16.2} & 18.6 & \textbf{9.1} & \textbf{12.2}  \\
\bottomrule
\end{tabular}
\end{adjustbox}
\label{tb:pcn}
\end{table}

\begin{figure}
\centering
\includegraphics[width=\columnwidth]{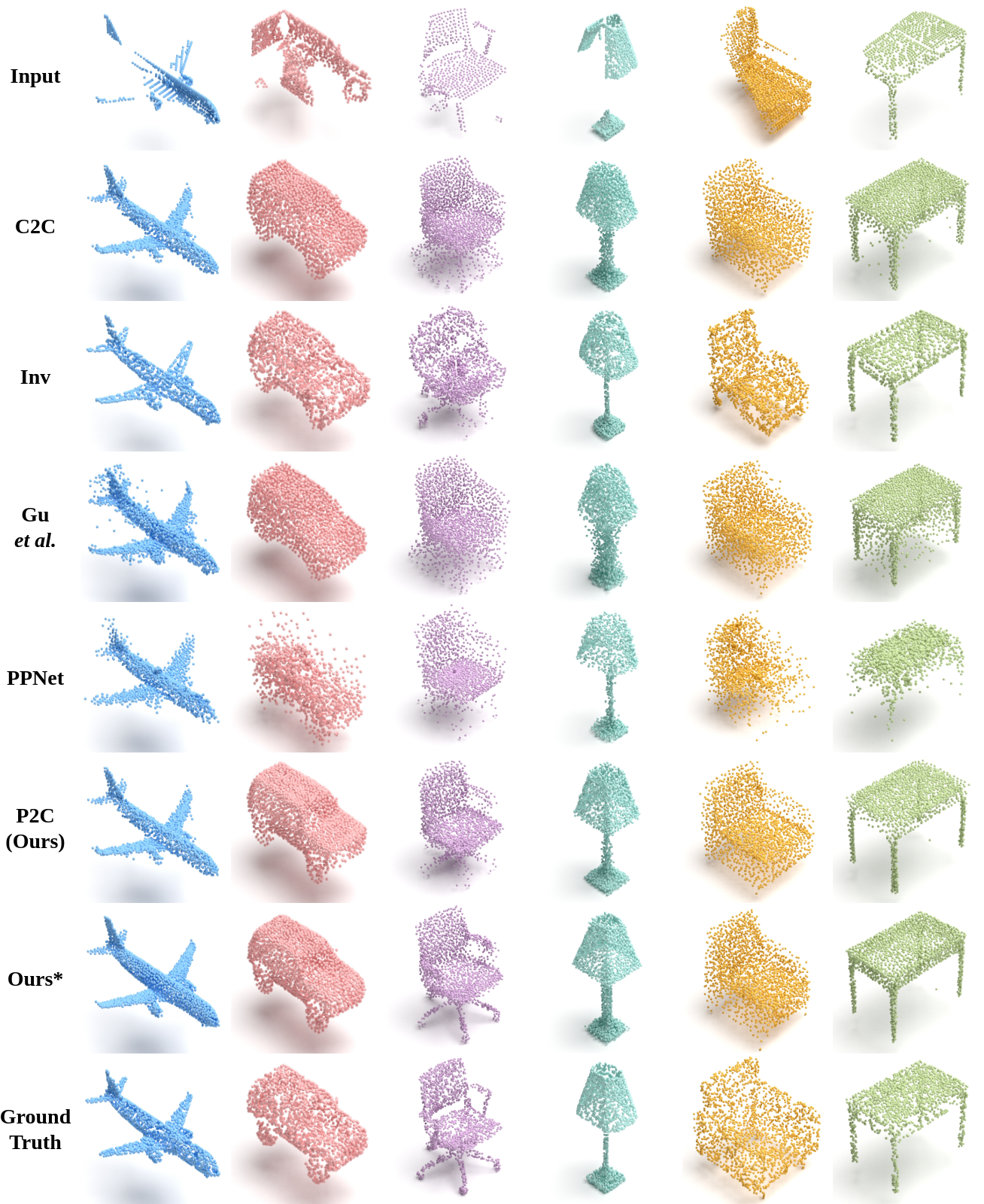}
\caption{Visual comparison of point cloud completion results on the 3D-EPN dataset.
}
\label{fig:epn3d}
\end{figure}

\noindent\textbf{Dataset.}
For a comprehensive comparison, we evaluate our method on synthetic and real-world datasets following state-of-the-art point cloud completion works~\cite{pointr_yu, cycle_wen}.
We evaluate our method on synthetic datasets 3D-EPN~\cite{epn_dai} and PCN~\cite{pcn_yuan}, where the former is usually adopted as an unpaired method benchmark, and the latter is widely used in supervised method evaluation.
Moreover, we also extract real-world objects from ScanNet~\cite{scannet_dai}.
In particular, 4357 chairs and 1271 tables are extracted as the training set, while the validation set contains 1368 chairs and 350 tables. ScanNet objects are unaligned and have around 800 points on average, creating a more challenging scenario.

\noindent\textbf{Evaluation Metric.}
We use $\ell_2$ Chamfer Distance (CD) as the evaluation metric for synthetic datasets. In the case of real-world datasets, where ground-truth complete shapes are unavailable, we evaluate the prediction in terms of both fidelity and quality.
To measure the preservation of observed regions in the prediction, we adopt the Unidirectional Chamfer Distance (UCD), Unidirectional Hausdorff Distance (UHD), and our proposed Region-Aware Chamfer Distance (RCD).
To evaluate the quality of the generated shapes, we utilize a complete shape example set extracted from ShapeNet~\cite{shapenet_chang} and employ the Minimal Matching Distance (MMD)~\cite{pointr_yu} as the quality metric.

\subsection{Synthetic Data Evaluation}

\begin{figure*}
\centering
\includegraphics[width=\textwidth]{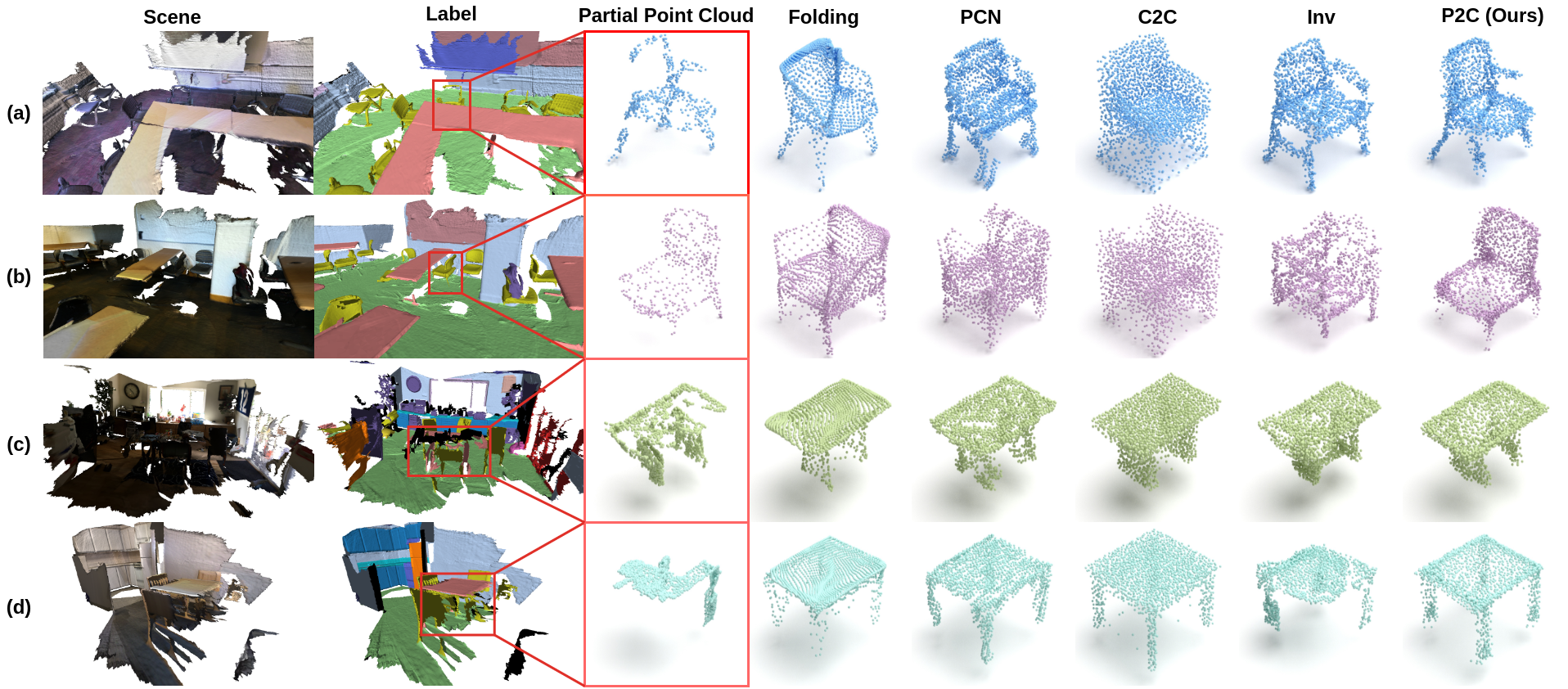}
\caption{Visual comparison of point cloud completion results on the ScanNet dataset.}
\label{fig:scan}
\end{figure*}

We compare the performance of our proposed P2C with state-of-the-art methods in the field, including supervised, unpaired (or unsupervised), and weakly-supervised methods.
To ensure a fair comparison, we use their open-source implementation and the same hyperparameters except Cai~\emph{et al}.~\cite{latent_cai} where open-source implementation is not available, so we cited their reported result.
Since the unsupervised method utilizes unpaired partial and complete samples for training, we provide results of our P2C trained with the same data source, indicated as P2C$^*$.
The results on the 3D-EPN dataset are shown in Tab.~\ref{tb:epn}, demonstrating the superiority of our method.
P2C outperforms the best unpaired method~\cite{latent_cai} by 2.7 w.r.t CD-$\ell_2$ without any design to utilize known complete example shapes.
Moreover, compared with the best weakly-supervised method~\cite{wild_gu}, our proposed P2C improves the CD score by 7.2 with only single partial observations for training.
Although fully supervised methods still show numerical advantages by heavily exploiting complete and paired ground-truth data, our self-supervised framework P2C has significantly reduced the performance gap between the two different learning schemes.

Tab.~\ref{tb:pcn} shows the performance comparison on the PCN dataset.
Our method is trained with the same data source as unpaired methods for a fair comparison.
On average, we achieve 12.2 CD-$\ell_2$, while other unpaired methods have around 14, showing that our approach attains a much better overall object completion quality.
The per-category results demonstrate that our proposed method outperforms the best unpaired model in six out of all eight testing categories.

Fig.~\ref{fig:epn3d} presents a qualitative comparison between our method and some recent methods~\cite{cycle_wen, shapeinver_zhang, wild_gu, pointpncnet_mittal}, showcasing that our method can successfully complete objects with diverse missing regions even in the absence of complete samples. In particular, our method trained on unpaired data recovers not only realistic geometry, such as the lamp post, but also captures fine-grained details, such as the car's wheel and the desk's edges.

\subsection{Real-world Data Evaluation}
\begin{table}
\centering
\caption{Shape completion comparison with supervised and unpaired methods on the ScanNet dataset. The numbers shown are RCD $\downarrow$, UCD-$\ell_2$ $\downarrow$, UHD $\downarrow$, and MMD $\downarrow$ scaled by $10^3$ and $10^4$, $10^2$, and $10^3$, respectively.}
\begin{adjustbox}{width=\columnwidth, center}
\setlength\tabcolsep{2pt}
\begin{tabular}{l|cc|cc|cc|cc}
\toprule
Metric        & \multicolumn{2}{c|}{RCD}                               & \multicolumn{2}{c|}{UCD}                               & \multicolumn{2}{c|}{UHD}                               & \multicolumn{2}{c}{MMD}                               \\ \midrule
Category      & \multicolumn{1}{c}{Chair} & \multicolumn{1}{c|}{Table} & \multicolumn{1}{c}{Chair} & \multicolumn{1}{c|}{Table} & \multicolumn{1}{c}{Chair} & \multicolumn{1}{c|}{Table} & \multicolumn{1}{c}{Chair} & \multicolumn{1}{c}{Table} \\ \midrule
Folding~\cite{folding_yang}     & 14.2  & 11.9  & 124.6  & 86.1  & 23.5  & 16.9  & 6.5  & 8.0  \\
PCN~\cite{pcn_yuan}             & 17.9  & 14.9  & 131.8  & 85.1  & 24.5  & 16.8  & 5.9  & 7.2  \\
TopNet~\cite{topnet_Tchapm}     & 20.3  & 14.1  & 114.6  & 82.5  & 23.0  & 16.7  & 5.8  & 7.6  \\ \midrule
C2C~\cite{cycle_wen}            & 16.2  & 10.1  & 18.5   & 14.6  & 13.0  & 10.2  & \bt{9.8}  & 9.1  \\
Inv~\cite{shapeinver_zhang}     & 18.4  & 9.5   & 8.5    & 7.5   & 10.0  & 8.6   & 15.2 & 16.2 \\
P2C(Ours)                       & \bt{4.6}  & \bt{6.7}  & \bt{7.7}  & \bt{7.2}  & \bt{8.3}  & \bt{8.2}  & 14.1  & \bt{8.1}  \\ \bottomrule
\end{tabular}

\end{adjustbox}
\label{tb:scan}
\end{table}

We evaluate the effectiveness of our method on the ScanNet dataset by training P2C on only partial objects and comparing it with relevant methods pretrained on the ShapeNet dataset.
The results are shown in Tab.~\ref{tb:scan}, which indicates that our method outperforms methods trained with complete shape examples in terms of fidelity (RCD, UCD, UHD), including both supervised and unsupervised methods.
While we achieve the best result in the table category compared with other unpaired methods using MMD as a quality measure, the unpaired method C2C~\cite{cycle_wen} outperforms ours on the chair category and supervised methods perform better than ours on both categories. Considering the fact that MMD measures the distance of a prediction and its ShapeNet nearest neighbor~\cite{pointr_yu}, the MMD scores for the compared methods are usually better since they are all trained on ShapeNet to closely resemble ShapeNet samples.

\subsection{Ablation Study}
\begin{table}
\centering
\caption{Ablation study on four categories (plane, car, chair, table) of the 3D-EPN dataset. We investigate the impact of RCD, $\mathcal{L}_f$, and NCC designs. Results reported in CD-$\ell_2$ $\downarrow$ scaled by $10^4$.}
\begin{adjustbox}{width=0.80\columnwidth,center}
\begin{tabular}{c|ccc|a}
\toprule
Model & RCD           & Latent Recon. &    NCC      &  CD-$\ell_2$ \\ \midrule
A     &               &              &               &  18.6        \\
B     & \checkmark    &              &               &  13.5        \\
C     & \checkmark    & \checkmark   &               &  12.0        \\ \midrule
D     & \checkmark    & \checkmark   & \checkmark    &  \textbf{11.2}        \\ 
\bottomrule
\end{tabular}
\end{adjustbox}
\label{tb:part}
\end{table}
\noindent\textbf{Model Design Analysis.}
To examine the effectiveness of our design, we conducted a detailed ablation study on the key components using four main categories in the 3D-EPN dataset.
The Chamfer Distance-based evaluation results are summarized in Tab.~\ref{tb:part}.
The baseline model (Model A) is the same framework, employing only reconstruction and completion losses.
We then replace the vanilla CD measure used in the baseline model with the proposed RCD in Eq.~\ref{eq:rcd} to form Model B and observe a significant improvement compared to the baseline.
This is because the vanilla CD restricts the prediction to overfit existing regions, thereby preventing the model from inferring missing regions.
When the latent reconstruction loss ($\mathcal{L}_{f}$) is incorporated in Model C, the performance increases by 1.5 compared to Model B, indicating the effectiveness of $\mathcal{L}_{f}$.
Finally, to retain more completeness but fewer outliers, we further introduce the NCC (Eq.~\ref{eq:ncc}) to form our complete P2C framework (Model D), which helps to establish a state-of-the-art result as shown in Tab.~\ref{tb:epn}.

\noindent\textbf{Region-Aware Chamfer Distance.}
\begin{figure}
\centering
\includegraphics[width=0.98\columnwidth]{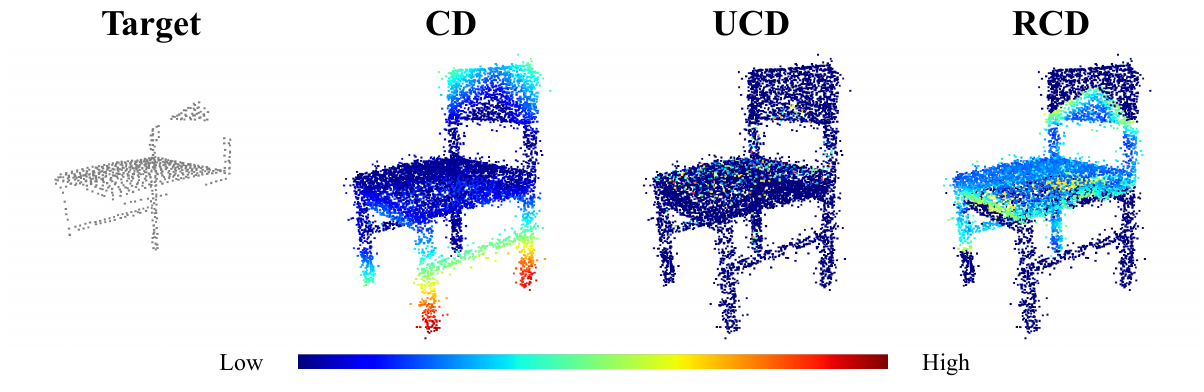}
\caption{Visualization of per-point distance measures.
Our proposed RCD computes point cloud distance by considering the corresponding regions that appear in the target shape.
In contrast, the vanilla CD assigns large distances to the unseen parts of the candidate shape, while the UCD lacks a distance measure to outliers near the observed part of the target shape.}
\label{fig:rcd_vis}
\end{figure}
We demonstrate the effect of RCD in Fig.~\ref{fig:rcd_vis}, where we visualize different distance measures.
Given a partial shape and a noisy complete prediction, vanilla CD computes the distances for all points in the prediction without considering some parts that have no correspondence in the partial observation.
This issue causes substantial distance estimations for missing parts.
Consequently, as the objective function aims to minimize distance, an overfitted network will reconstruct the exact inputs instead of recovering missing regions.
On the other hand, UCD lacks regularization to outlier points as those points are usually further from observed points than valid point predictions. This limitation allows the network to cheat the metric by outputting a shape that fills the whole 3D space, in which case the UCD value will be zero for any possible input shape.
Our proposed RCD addresses the above problems by introducing region awareness since we only evaluate distance for points near the observed region, assigning no distance to unseen parts. Therefore, outlier points are constrained in the observed region, while completion capability will not be restricted.

\begin{table}
\centering
\caption{The effect of different schemes for enforcing local planarity in CD-$\ell_2$ $\downarrow$ scaled by $10^4$.}
\begin{adjustbox}{width=0.85\columnwidth,center}
\begin{tabular}{l|a|cccc}
\toprule
Method         & Average & Plane & Car  & Chair & Table \\ \midrule
Baseline       & 13.5    & 4.7   & 14.2 & 14.4  & 20.8  \\
Mean           & 12.8    & 4.4   & 12.3 & 14.2  & 19.8  \\ 
Min            & 13.2    & 4.7   & 13.5 & 14.4  & 20.1  \\ \midrule
Variance       & \bt{11.6} & \bt{4.3}   & \bt{10.1}  & \bt{13.9}  & \bt{18.2}  \\ 
\bottomrule
\end{tabular}
\end{adjustbox}
\label{tb:ncc}
\end{table}
\noindent\textbf{Normal Consistency Constraint.}
To evaluate the effectiveness of NCC in improving surface continuity, we compare two alternative strategies for calculating the normal consistency (Eq.~\ref{eq:nc}) of a given point. Instead of incorporating the local planarity, we estimate the curvature as the mean of normal vector dot product similarity or minimum of the similarity and compare our method with them. Tab.~\ref{tb:ncc} shows the quantitative results comparing the strategies, where the baseline model is a simplified variant that only utilizes the reconstruction and completion losses. Based on the mean similarity and minimal similarity, we observe incremental improvements compared to the baseline model, where the average CD-$\ell_2$ only drops from 13.5 to 12.8 and 13.2, respectively. In comparison, our proposed NCC that uses the variance can better estimate the local surface curvature and improve the completion quality.

We provide more ablation studies in the supplementary material, including model complexity and efficiency, empirical hyperparameter selection, visualizations, \emph{etc}.


\section{Conclusion}

In this paper, we propose P2C, the first self-supervised point cloud completion method that only requires a single partial point cloud observation per object for learning. Our method employs a novel Region-Aware Chamfer Distance to measure input-prediction similarity, and we design the Normal Consistency Constraint to enhance prediction completeness. Experimental results demonstrate that P2C exhibits state-of-the-art performance on both synthetic and real-world completion tasks, even outperforming models trained with known complete samples. Overall, our proposed method provides an effective solution for point cloud completion given only partial observation data.

\clearpage
\newpage
{\small
\bibliographystyle{ieee_fullname}
\bibliography{reference}
}

\end{document}